\documentclass{article}
\usepackage{iclr2026_conference,times}

\setcitestyle{numbers,sort&compress}

\usepackage{amsmath,amsfonts,bm}

\def\eqref#1{equation~\ref{#1}}

\def\1{\bm{1}}

\DeclareMathAlphabet{\mathsfit}{\encodingdefault}{\sfdefault}{m}{sl}
\SetMathAlphabet{\mathsfit}{bold}{\encodingdefault}{\sfdefault}{bx}{n}

\usepackage{booktabs}
\usepackage{graphicx}
\usepackage{url}
\usepackage{hyperref}
\usepackage{cleveref}

\title{Impermanent: A Live Benchmark for Temporal Generalization in Time Series Forecasting}

\author{Anonymous Authors \\
Paper under double-blind review.
}
\author{
\mbox{}\And
Azul Garza \quad  \quad Ren\'ee Rosillo \\
TimeCopilot \\
\texttt{\{azul,renee\}@timecopilot.dev}
\And\mbox{}
\AND
Rodrigo Mendoza-Smith \\
Independent Researcher
\And
David Salinas \\
ELLIS Institute T\"ubingen
\And
Andrew Robert Williams \\
Mila -- Quebec AI Institute \\
Universit\'e de Montr\'eal
\AND
Arjun Ashok \\
Mila -- Quebec AI Institute \\
Universit\'e de Montr\'eal
\And
Mononito Goswami \\
Amazon Web Services\textsuperscript{\dag}
\And
Jos\'e Mart\'in Ju\'arez\\
TimeCopilot
}

\iclrfinalcopy

\begin{document}

\maketitle

\begingroup
\renewcommand\thefootnote{}
\footnotetext{$^{\dag}$ The work does not relate to the author’s position at Amazon.}
\endgroup

\newcommand{\Impermanent}{{\tt Impermanent}}

\begin{abstract}
    Recent advances in time-series forecasting increasingly rely on  pre-trained foundation-style models. While these models often claim broad generalization, existing evaluation protocols provide limited evidence. Indeed, most current benchmarks use static train-test splits that can easily lead to contamination as foundation models can inadvertently train on test data or perform model selection using test scores, which can inflate performance.
    We introduce \Impermanent{}, a live benchmark that evaluates forecasting models under open-world temporal change by scoring forecasts sequentially over time on continuously updated data streams, enabling the study of temporal robustness, distributional shift, and performance stability rather than one-off accuracy on a frozen test set. 
    \Impermanent{} is instantiated on GitHub open-source activity, providing a naturally live and highly non-stationary dataset shaped by releases, shifting contributor behavior, platform/tooling changes, and external events.
    We focus on the top 400 repositories by star count and construct time series from issues opened, pull requests opened, push events, and new stargazers, evaluated over a rolling window with daily updates, alongside standardized protocols and leaderboards for reproducible, ongoing comparison.
    By shifting evaluation from static accuracy to sustained performance, \Impermanent{} takes a concrete step toward assessing when—and whether—foundation-level generalization in time-series forecasting can be meaningfully claimed. Code and a live dashboard are available at \url{https://github.com/TimeCopilot/impermanent} and \url{https://impermanent.timecopilot.dev}.
\end{abstract}

\section{Introduction}

Forecasting seeks to reduce uncertainty about the future.
In practice, forecasters rely on a diverse set of models (statistical methods, machine learning approaches, and increasingly, foundation models), selecting among them based on properties of the data such as trend, seasonality, sparsity, scale, and domain structure, and validating choices through backtesting on historical observations \cite{hyndman2021fpp3}.
In recent years, \emph{time-series foundation models} (TSFMs) have emerged with the promise of broad generalization across domains \citep{ekambaram2024tinytimemixersttms,goswami2024momentfamilyopentimeseries,woo2024unifiedtraininguniversaltime,liu2025timerxllongcontexttransformersunified,shi2025timemoebillionscaletimeseries,cohen2024tototimeseriesoptimized,liu2024timergenerativepretrainedtransformers,das2023decoder,rasul2023lag,ansari2024chronoslearninglanguagetime,liu2024autotimesautoregressivetimeseries,gruver2024largelanguagemodelszeroshot,jin2023time,lu2021pretrainedtransformersuniversalcomputation,garza2023timegpt}.
Trained on large and heterogeneous collections of temporal data, these models aim to operate as generalists, producing accurate forecasts on previously unseen series with minimal adaptation.
The central claim underlying this paradigm is \emph{temporal generalization}, i.e. that learned representations transfer across datasets, frequencies, and domains.

However, despite this promise, most empirical evaluations of TSFMs rely on \emph{static} benchmarks. 
Datasets are fixed in time, and evaluation is conducted on held-out splits drawn from the same underlying distribution as the training data, as in widely used benchmarks such as GIFT-Eval\cite{aksu2024gifteval}, FEV \cite{shchur2025fevbench}, and the Monash Forecasting Repository\cite{godahewa2021monash}.
While this protocol measures cross-sectional generalization, it does not fully test whether performance persists \emph{across time} in evolving, non-stationary environments.
Real-world forecasting is inherently dynamic: data distributions shift, new series appear, and structural breaks occur.
Moreover, public datasets used for pretraining are often reused in downstream benchmarks, raising concerns about memorization, data leakage, and test-set contamination \cite{meyer2025time}.
These concerns become increasingly relevant as foundation models grow in scale and as training data curation becomes less transparent.
%

The importance of this sequential-evaluation requirement is well recognized in adjacent literatures.
For example, the \emph{prequential} viewpoint holds that probabilistic forecasts should be evaluated in the order they are made \cite{dawid1984prequential}.
More broadly, the concept-drift literature examines how predictive systems degrade as data distributions evolve over time \cite{gama2014conceptdrift}.
In forecasting practice, rolling-origin evaluation is similarly recommended because it better reflects real-world deployment \citep{Armstrong01, hyndman2024fpppy}.
Related concerns also arise for language foundation models, where live benchmarks aim to reduce contamination and ensure that reported results are not affected by prior exposure to test examples \cite{white2024livebench,chiangChatbotArenaOpen2024}.
Closest to our setting, \cite{kargerForecastBenchDynamicBenchmark2025} considers sequential evaluation for LLM prediction of single future events, but not for time-series forecasting.

\paragraph{Contributions.}
To address these limitations, we introduce \Impermanent{}, which, to our knowledge, is the first \emph{live} benchmark designed specifically to evaluate \emph{temporal generalization} in time-series forecasting.
\Impermanent{} makes temporal generalization measurable through a live, leak-proof evaluation protocol in which forecasts are generated and scored sequentially over time.
This setup enables analyses that are difficult in static benchmarks, including sustained accuracy, robustness to distribution shift and shocks, and the stability of model rankings under ongoing change.
This first iteration of \Impermanent{} is built on GitHub software development activity, a naturally live and highly non-stationary environment shaped by releases, shifting contributor behavior, workflow and tooling changes, and external events \cite{gousios2012ghtorrent,gharchive,decan2020gap}. 
The benchmark is designed around a prequential, deployment-faithful evaluation loop: at each cutoff time, models must produce forecasts \emph{before} the corresponding ground truth exists.

\section{The \Impermanent{} Benchmark}

\begin{figure*}[t]
    \centering
    \includegraphics[width=0.85\textwidth]{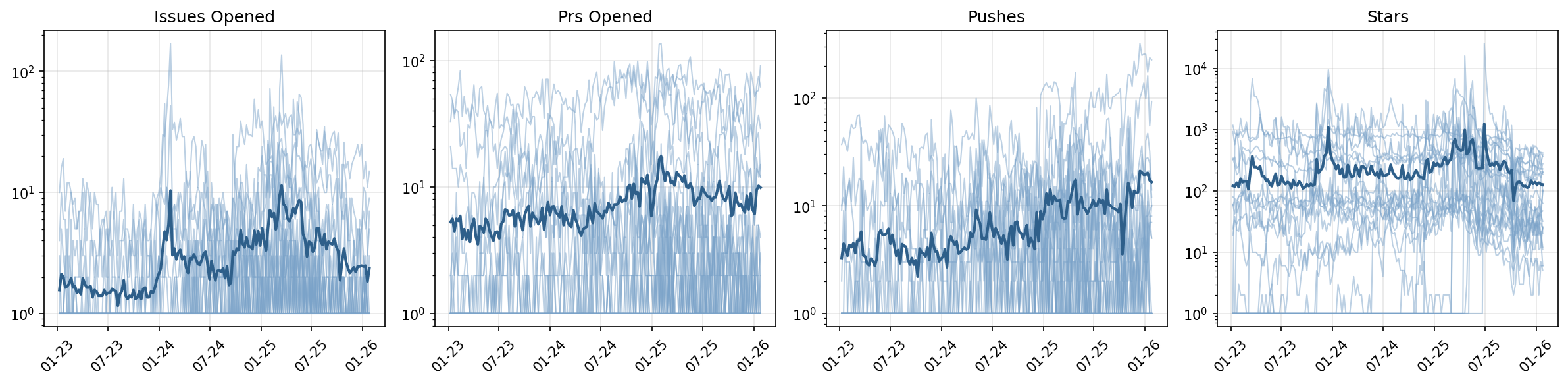}
    \vspace{2mm}
    \includegraphics[width=0.85\textwidth]{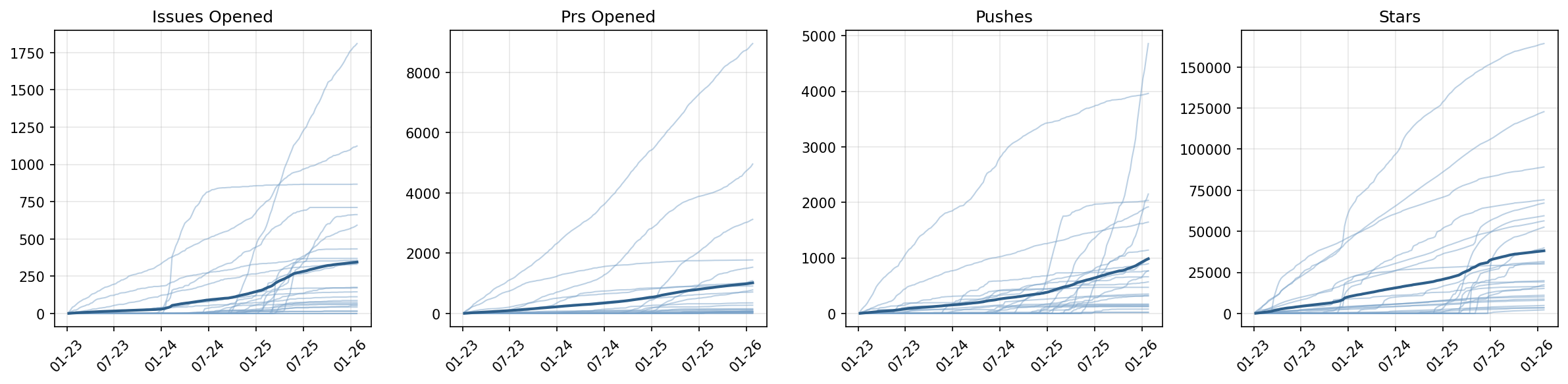}
    \caption{Weekly GitHub activity for a random sample of 25 repositories across four event types (issues opened, pull requests opened, push events, new stargazers). Top: weekly event counts. Bottom: cumulative counts over time.}
    \label{fig:weekly-by-event}
\end{figure*}

\begin{figure*}[t]
    \centering
    \includegraphics[width=0.95\textwidth]{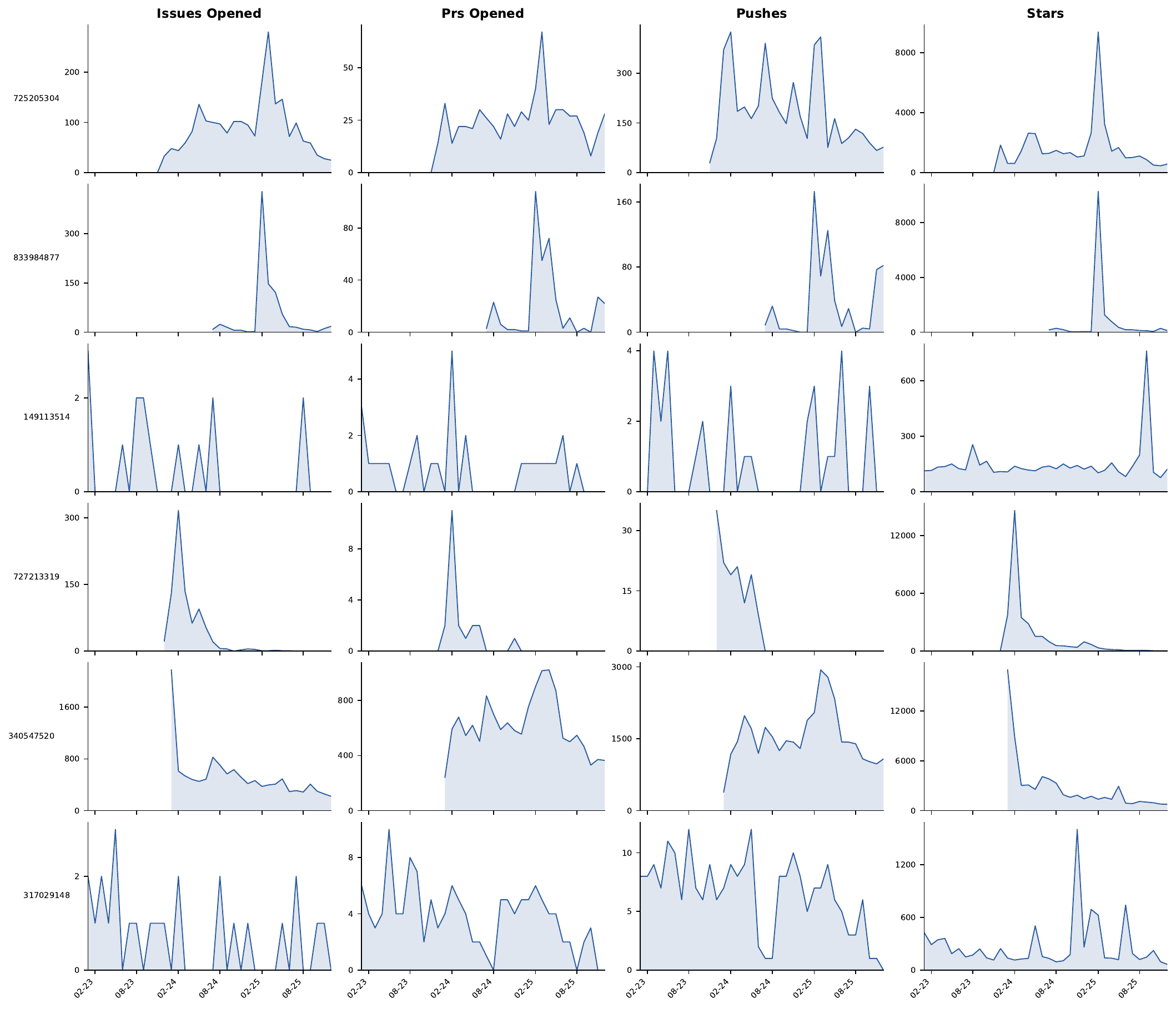}
    \caption{Monthly event counts for a random sample of six repositories across four activity signals.  The series exhibit hallmark features of real-world temporal data: intermittency (long zero-count stretches), burstiness (isolated spikes), level shifts, and heterogeneous scales across repositories and event types.}
    \label{fig:series-panel}
\end{figure*}

We instantiate \Impermanent{} on GitHub activity using GH Archive event streams \cite{gharchive} (\Cref{fig:series-panel}). We track four event types (issues opened, pull requests opened, push events, and new stargazers) for 400 repositories (selected by star count), construct time series at four forecast frequencies (hourly, $h=24$; daily, $h=7$; weekly, $h=4$; and monthly, $h=1$), and stratify repositories by activity level.In the current release, each repository--event-type pair defines a \emph{univariate} series (we forecast one target count at a time); capturing cross-repository co-movement and incorporating additional covariates are important next steps. Figures~\ref{fig:weekly-by-event} and \ref{fig:weekly-stats} provide exploratory views on a random sample of 25 repositories, visualizing non-stationarity in raw weekly counts and in compact dynamics descriptors, including the \emph{spectral centroid} and \emph{spectral entropy}. Formally, let $y_0,\ldots,y_{n-1}$ denote a series, $X_k=\sum_{t=0}^{n-1} y_t e^{-i2\pi kt/n}$ its real-FFT coefficients for $k=0,\ldots,K-1$ with $K=\lfloor n/2\rfloor+1$, and $f_k=k/n$ (cycles per time step); defining $P_k=|X_k|^2$ and $p_k=P_k/\sum_j P_j$, we compute the spectral centroid $C$ and normalized spectral entropy $H$ as:
\begin{equation}
    C = \frac{\sum_{k=0}^{K-1} f_k\, |X_k|}{\sum_{k=0}^{K-1} |X_k|},\qquad H = -\frac{1}{\log K}\sum_{k=0}^{K-1} p_k \log p_k.
\end{equation}
Qualitatively, larger $C$ indicates relatively faster dynamics (more power at higher frequencies), while larger $H$ indicates a more diffuse, less structured spectrum (power spread across many frequencies). In Figure~\ref{fig:weekly-stats}, we see that the GitHub streams occupy a wide range of $(C,H)$ values and that different event types populate different regions of this space: some series concentrate power at lower frequencies (lower $C$) with more structured spectra (lower $H$), while others show higher-frequency, more broadband behavior (higher $C$ and/or higher $H$). For \Impermanent{}, the key takeaway is simple: the dataset mixes smooth, trend-like behavior with spiky, volatile behavior, so a good forecaster must handle both slow changes and sudden bursts rather than fitting one fixed pattern.

\section{Evaluation Protocol}

At each cutoff date, every model receives a context window of historical observations and must produce point and probabilistic forecasts for the next $h$ periods before any ground truth is available.  Cutoffs are spaced exactly one horizon apart, and the most recent cutoff is always excluded because observations may still be incomplete.  \Cref{tab:eval-protocol} summarizes the evalution protocol.
\begin{table}[h!]
\centering
\begin{tabular}{@{}lcccc@{}}
\toprule
 & \textbf{Hourly} & \textbf{Daily} & \textbf{Weekly} & \textbf{Monthly} \\
\midrule
Horizon $h$          & 24      & 7     & 1      & 1      \\
Max context          & 1\,024  & 512   & 114    & 24     \\
Cutoff step          & 24\,h   & 7\,d  & 1\,w   & 1\,mo  \\
First cutoff         & 2026-02-08 & 2026-01-04 & 2026-01-04 & 2025-10-01 \\
\bottomrule
\end{tabular}
\caption{Evaluation protocol parameters. All frequencies use MASE and scaled CRPS.}
\label{tab:eval-protocol}
\end{table}

We evaluate forecasts with two complementary metrics: MASE (Mean Absolute Scaled Error) for point accuracy \cite{HK06}, and scaled CRPS for the predictive distribution \cite{Gneiting2014}, estimated from nine quantile levels $\mathcal{Q}=\{0.1,0.2,\ldots,0.9\}$. For a series with training window $y_{1:T}$, seasonal period $m$, and horizon $h$, we use
\begin{equation}
\mathrm{MASE}=\frac{\frac{1}{h}\sum_{i=1}^{h}|y_{T+i}-\hat{y}_{T+i}|}{\frac{1}{T-m}\sum_{t=m+1}^{T}|y_t-y_{t-m}|},\qquad
\mathrm{CRPS}\approx\frac{1}{h}\sum_{i=1}^{h}\frac{2}{|\mathcal{Q}|}\sum_{\tau\in\mathcal{Q}}\rho_\tau\!\left(y_{T+i}-\hat{q}_{T+i}(\tau)\right),
\end{equation}
where $\rho_\tau(u)=u\bigl(\tau-\mathbf{1}\{u<0\}\bigr)$ is the pinball loss. We compute both metrics per series and aggregate per subdataset using the median. To make scores comparable across subdatasets, we scale each metric by the \emph{zero model}, which always predicts zero: for model value $v$ and zero-model score $b$ (the metric value from all-zero forecasts), we report $v/\max(b,\tau_0)$, where $\tau_0$ is the 10th percentile of strictly positive zero-model scores for that metric. This floor prevents unstable ratios when $b$ is very small.
We evaluate eleven models in three groups. Baselines are \emph{ZeroModel} (which always predicts zero and provides the scaling denominator), \emph{HistoricAverage}, and \emph{SeasonalNaive}. Statistical models are AutoARIMA~\cite{HK08}, AutoETS~\cite{HKSG02}, AutoCES \cite{sventukovces}, Dynamic Optimized Theta~\cite{FIORUCCI20161151}, and Prophet~\cite{prophet}, and all run on CPU. Foundation models are Chronos-2~\cite{ansari2025chronos2univariateuniversalforecasting}, Moirai~2.0-R-Small~\cite{woo2024unifiedtraininguniversaltime}, TimesFM~2.5~\cite{das2023decoder}, and TiRex~\cite{auer2025tirexzeroshotforecastinglong}, and each runs on an A10G GPU with batch size 64. Each model outputs nine quantile forecasts ($\tau \in \mathcal{Q}$), enabling direct comparison on MASE and scaled CRPS. We include only open source TSFMs with released weights and reproducible inference code, and we run all models through TimeCopilot~\cite{garza2025timecopilot}. \Cref{sec:infrastructure} describes the pipeline.

\section{Results}

\Cref{tab:leaderboard} illustrates the kind of analysis \Impermanent{} enables.  In this early snapshot up to February 12th, 2026, pre-trained foundation models occupy the top four positions, with TimesFM leading on three of four columns.  However, the picture is nuanced: SeasonalNa\"{i}ve achieves a competitive MASE rank (5.39) while showing poor probabilistic calibration (CRPS rank 9.50), and AutoETS and AutoARIMA attain CRPS ranks comparable to DynOptTheta despite weaker point accuracy.  Because \Impermanent{} scores models sequentially over an evolving stream, these rankings will shift as new cutoffs accumulate, making it possible to track \emph{whether} early advantages persist under continued distributional shift rather than taking any one leaderboard snapshot as definitive.

\begin{table}[h!]
\centering
\begin{tabular}{@{}lcccc@{}}
\toprule
 & \multicolumn{2}{c}{Median value $\downarrow$} & \multicolumn{2}{c}{Mean rank $\downarrow$} \\
\cmidrule(lr){2-3} \cmidrule(lr){4-5}
\textbf{Model} & MASE & CRPS & MASE & CRPS \\
\midrule
HistoricAverage        & 4.740          & 3.669          & 9.943          & 8.401          \\
SeasonalNaive      & 1.272          & 2.950          & 5.385          & 9.495          \\
\midrule
Prophet~\cite{prophet} & 4.264          & 6.713          & 9.791          & 8.638          \\
AutoCES~\cite{sventukovces}                & 2.272          & 2.385          & 7.293          & 6.433          \\
AutoARIMA~\cite{HK08} & 3.157          & 2.258          & 7.842          & 5.840          \\
AutoETS~\cite{HKSG02} & 2.802          & 2.232          & 7.119          & 5.864          \\
DynOptTheta~\cite{FIORUCCI20161151} & 1.522          & 2.494          & 5.838          & 6.088          \\
\midrule
Chronos~\cite{ansari2025chronos2univariateuniversalforecasting} & 0.789          & 2.341          & 3.340          & 4.348          \\
Moirai~\cite{woo2024unifiedtraininguniversaltime} & 0.786          & \textit{2.153} & 3.028          & \textit{4.173} \\
TiRex~\cite{auer2025tirexzeroshotforecastinglong} & \textit{0.757} & 2.270          & \textbf{2.938} & 4.223          \\
TimesFM~\cite{das2023decoder} & \textbf{0.609} & \textbf{1.055} & \textit{2.979} & \textbf{2.041} \\
\bottomrule
\end{tabular}
\caption{Overall leaderboard results on \Impermanent{}, aggregated across all subdatasets, frequencies, and cutoff dates.  \emph{Median value} reports the median of the baseline-scaled metric across all evaluation instances; \emph{mean rank} is the average rank computed hierarchically (within each group, then across subdatasets and frequencies).  \textbf{Bold}: best per column; \textit{italic}: second best.  Models are sorted by the average of the two rank columns (worst first); horizontal rules separate naive baselines, statistical methods, and foundation models.}
\label{tab:leaderboard}
\end{table}

\section{Conclusion and Future Work}

We introduced \Impermanent{}, which, to our knowledge, is the first live benchmark designed to measure temporal generalization in time-series forecasting. By evaluating models sequentially on a continuously evolving data stream, with forecasts issued before outcomes are observed and scored only after those outcomes arrive, \Impermanent{} enables analyses that static benchmarks cannot support. These include measuring sustained accuracy over time, assessing robustness under distributional change, and tracking the stability of model rankings as evaluation unfolds. All data pipelines, evaluation code, and leaderboard infrastructure are open and fully automated, enabling reproducibility and extension without reprocessing historical data.

This first iteration of \Impermanent{} is built on GitHub software development activity, but the framework is designed to support broader future development. Natural next steps include expanding to additional live data streams, enriching forecasting tasks with auxiliary contextual information, and using longer evaluation horizons to better understand performance stability and ranking dynamics over time. More broadly, we hope \Impermanent{} can serve as a shared resource for studying whether benchmark performance in static settings translates into reliable performance after deployment.

\pagebreak

\bibliographystyle{unsrtnat}
\bibliography{refs}

\newpage
\section*{Appendix}

\subsection*{Exploratory Statistics}
\begin{figure*}[h!]
    \centering
        \caption{Weekly-series summary statistics for a random sample of 25 repositories, grouped by event type (columns). Each panel shows a relationship between two descriptors: log low/high bandpower ratio (slow vs fast variation) vs permutation entropy (irregularity), and spectral centroid (typical frequency) vs spectral entropy (spectral spread).}
    \label{fig:weekly-stats}
    \includegraphics[width=0.98\textwidth]{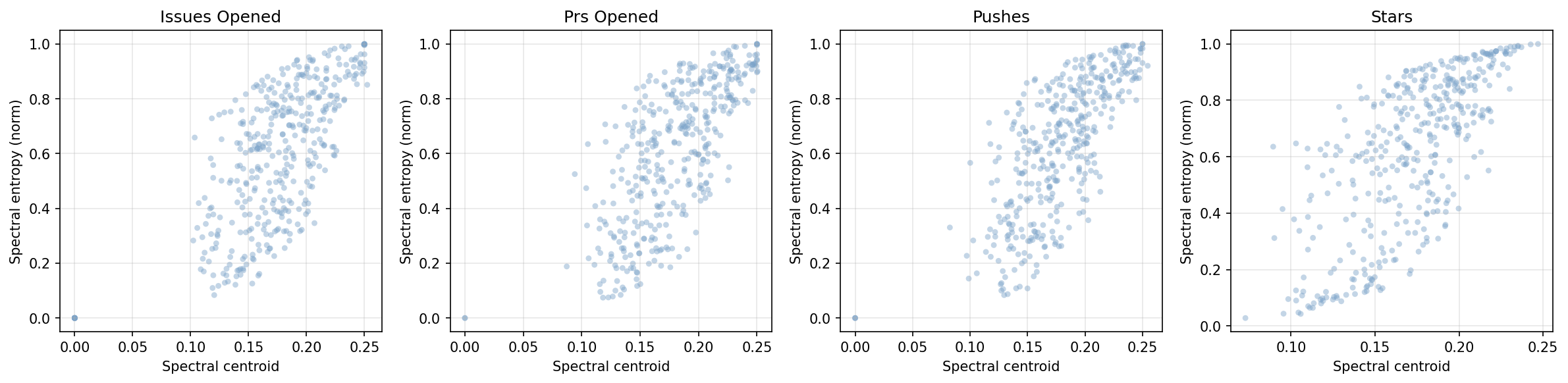}
\end{figure*}

This section provides additional descriptive views of the GitHub activity streams used in \Impermanent{}, complementing the main-text plots in Figures~\ref{fig:weekly-by-event} and~\ref{fig:weekly-stats}. The weekly trajectories reveal pronounced intermittency and burstiness: many repository--event series spend long stretches near low counts and then undergo short-lived spikes, especially for issues and pushes, while stars operate at substantially larger scales with occasional surges. In cumulative form, these same bursts appear as slope changes and step-like jumps, indicating that activity regimes are not stable over time even within a single repository--event pair.

Figure~\ref{fig:weekly-stats} provides a compact summary through spectral descriptors. Across all four event types, spectral entropy generally increases with spectral centroid, producing an upward wedge-shaped cloud: faster series tend to exhibit broader-band spectra. At the same time, dispersion differs by event type: pull-request and push series are concentrated in mid-to-high centroid regions with moderate-to-high entropy, whereas stargazer series span a wider range from low-centroid/low-entropy to high-centroid/high-entropy regimes. This heterogeneity reinforces the motivation for \Impermanent{}: evaluation should track performance sequentially over time rather than rely on a single static slice.

\subsection*{Infrastructure} \label{sec:infrastructure}

\Impermanent{} runs as a set of serverless pipelines on Modal, with artefacts stored on Amazon~S3.  The pipeline has three stages.  \emph{Data ingestion} downloads hourly JSON archives from GH~Archive, filters for the four target event types, aggregates per-repository counts with DuckDB, and rolls them up to the daily, weekly, and monthly granularities (subject to completeness thresholds of 90\%, 95\%, and 99\% of constituent hours, respectively).  \emph{Forecasting} dispatches one job per \texttt{(model, cutoff)} pair: CPU models run on 32-core instances, foundation models on NVIDIA A10G GPUs, with up to 125 containers in parallel.  \emph{Evaluation} scores stored forecasts once ground truth arrives and rebuilds the leaderboard by reading all per-cutoff metric files, scaling by the zero-model baseline, and writing a single result Parquet.  The full cycle is triggered weekly; every stage is idempotent, so re-runs skip completed work and new models can be added without reprocessing history.

\end{document}